\documentclass[letterpaper]{article} 
\usepackage{aaai24}  
\usepackage{times}  
\usepackage{helvet}  
\usepackage{courier}  
\usepackage[hyphens]{url}  
\usepackage{graphicx} 
\usepackage{natbib}  
\usepackage{caption} 
\usepackage{algorithm}
\usepackage{algorithmic}

\usepackage{newfloat}
\usepackage{listings}
\DeclareCaptionStyle{ruled}{labelfont=normalfont,labelsep=colon,strut=off} 
\floatstyle{ruled}
\newfloat{listing}{tb}{lst}{}
\floatname{listing}{Listing}
\usepackage{amsmath}
\usepackage{amssymb}

\usepackage{booktabs} 
\usepackage{multirow}
\usepackage{color, colortbl}
\usepackage{adjustbox}
\usepackage{xspace}
\usepackage{amsfonts, subcaption}

\newcommand\norm[1]{\left\lVert#1\right\rVert}

\title{Few-Shot Object Detection via Synthetic Features with Optimal Transport}
\author {
    Anh-Khoa Nguyen Vu\textsuperscript{\rm 1,2}, 
    Thanh-Toan Do\textsuperscript{\rm 3}, 
    Vinh-Tiep Nguyen\textsuperscript{\rm 1, 2}, 
    Tam Le\textsuperscript{\rm 4}, 
    Minh-Triet Tran\textsuperscript{\rm 5, 2}, 
    Tam V. Nguyen\textsuperscript{\rm 6}
}
\affiliations {
    \textsuperscript{\rm 1}University of Information Technology, Ho Chi Minh City, Vietnam\\
    \textsuperscript{\rm 2}Vietnam National University, Ho Chi Minh City, Vietnam\\
    \textsuperscript{\rm 3}Department of Data Science and AI, Monash University, Australia\\
    \textsuperscript{\rm 4}The Institute of Statistical Mathematics / RIKEN AIP\\
    \textsuperscript{\rm 5}University of Science, Ho Chi Minh City, Vietnam\\
    \textsuperscript{\rm 6}University of Dayton, Ohio, U.S.\\
    khoanva@uit.edu.vn, 
    toan.do@monash.edu, 
    tiepnv@uit.edu.vn, 
    tam@ism.ac.jp,
    tmtriet@hcmus.edu.vn,
    tamnguyen@udayton.edu 
}

\usepackage{bibentry}

\begin{document}

\maketitle

\begin{abstract}


Few-shot object detection aims to simultaneously localize and classify the objects in an image with limited training samples. However, most existing few-shot object detection methods focus on extracting the features of a few samples of novel classes that lack diversity. Hence, they may not be sufficient to capture the data distribution. To address that limitation, in this paper, we propose a novel  approach in which we train a generator to generate synthetic data for novel classes. Still, directly training a generator on the novel class is not effective due to the lack of novel data. To overcome that issue, we leverage the large-scale dataset of base classes. Our overarching goal is to train a generator that captures the data variations of the base dataset. We then transform the captured variations into novel classes by generating synthetic data with the trained generator. To encourage the generator to capture data variations on base classes, we propose to train the generator with an optimal transport loss that minimizes the optimal transport distance between the distributions of real and synthetic data. Extensive experiments on two benchmark datasets demonstrate that the proposed method outperforms the state of the art. Source code will be available.

\end{abstract}
\section{Introduction}

Few-shot object detection (FSOD) consists of localizing the position of objects in images and predicting their labels given the limited labeled training data.
A common approach \cite{meta-rcnn, yolo-reweighting, tfa, defrcn, meta-detr} for FSOD is to train models on two sets of data, i.e., a base set and a novel set. 
A large base dataset with numerous training samples per base class is used to learn the concepts. Then, the detectors are fine-tuned on a small dataset containing novel classes with only few training samples per class. 

The trailblazer works \cite{meta-rcnn, yolo-reweighting} take two inputs, namely, query image and support image to detect novel objects. Those methods leverage the similarity of support and query features to determine instances in query images. Therefore, the subsequent methods \cite{max-margin, rpn-attention, han2022meta, trans-int} are introduced to fuse the cross-information between query and support features more effectively. On the other hand, other methods \cite{tfa, defrcn, zhu2021semantic} with only query images (i.e., without support images) also achieve competitive results. 


Nonetheless, the performance of  above methods in extremely scarce mode is far from satisfying because novel classes have little variety. The problem can be addressed by creating synthetic samples via leveraging the variation of base classes. A recent work \cite{zhang2021hallucination} deals with the FSOD via creating hallucination (synthetic) samples. 
In that work, the classifier and the generator are alternatively trained using multiple episodes of few-shot examples of novel classes. When training the generator in an episode the classifier is fixed and the classification error on the generated features is used to update the generator. Using multiple episodes of training with few-shot examples of novel classes could cause the model to overfit the training data of novel classes. 



To overcome the above issue, we propose a generative  approach for few-shot object detection via \textbf{S}ynthetic \textbf{F}eatures with \textbf{O}ptimal \textbf{T}rasport dubbed SFOT. Our method trains a generator to create synthetic features for novel classes. However, directly training a generator on novel class is ineffective due to the lack of novel data, which leads to overfitting. Thus, we leverage knowledge from the large-scale dataset by only training the generator on the base dataset to capture the data variation of base classes. We then transform the captured variations into novel classes by generating synthetic data with the trained generator during the fine-tuning stage. Through the use of the trained generator, we expect the data variation on the base classes can be simulated to novel classes.


To optimize the generator, we could use point-wise matching. However, it is sensitive to variations of samples, which is harmful to the generator training. Therefore, we rely on the distribution matching to train the generator that captures the variations of base classes. Specifically, the optimization of the matching between real and synthetic data is transformed into the optimal transport (OT) problem to obtain an effective generator. Still, increasing the number of synthetic features will lift the computational cost of the OT algorithm and raise the rate of noisy samples in the generated samples (e.g. a sample is in the dog class but its synthetic feature is close to the cat class). To overcome that issue, we integrate a clustering technique into OT method to cluster synthetic samples and then calculate the distribution distance between centroids and real samples. Moreover, this design allows us to form a distribution of synthetic samples based on the number of elements of each cluster, which is more flexible rather than the uniform distribution used in the previous work. To the best of our knowledge, this is the first work that leverages optimal transport for the few-shot object detection problem. As a result, our method creates synthetic novel features which improve the generalization of detectors in case of limited training data of novel classes.

Recently, \cite{wang2021zero} also uses optimal transport to minimize the distance between real and generated synthetic features. We would highlight three different points between our method and \cite{wang2021zero}. (1) Zero-shot learning methods usually use external knowledge (attributes, text) to extract the features of unseen classes, which is not essential for the FSOD methods because they can leverage a few images to describe a novel class in detail. Moreover, detectors need to localize the position of objects in novel classes, which creates more challenges to train an FSOD model. 
(2) In~\cite{wang2021zero}, Wang \textit{et al.} use a two-stage training, i.e., generator training and classifier training for classification. That training scheme in inapplicable to our FSOD context. Therefore, different from \cite{wang2021zero}, 
we propose a three-stage training, i.e., base class training, generator training, and few-shot fine-tuning. The three-stage is indispensable in our context because we need to guide the concepts of object detection. In addition, we also consider the imbalance between background and foreground objects during generator and classifier training. (3) We integrate a clustering method into the optimal transport algorithm to reduce the  computation cost of optimal transport and the effects of noisy samples. 


Our contributions are three-fold: (i) we propose a novel method to leverage variations on base classes to capture the distribution of novel classes through generating synthetic features; (ii) we effectively optimize the generator by solving the optimal transport problem between the real and synthetic feature distributions, which is integrated a clustering technique; 
(iii) we conduct extensive experiments and our SFOT method surpasses the state-of-the-art baselines on challenging FSOD benchmarks.








\section{Related Work}

\subsection{Object Detection} 
 
Deep learning-based object detection methods can be grouped into two main categories, i.e., two-stage detectors and one-stage detectors. Representatives of two-stage detectors are R-CNN methods, specifically Faster R-CNN~\cite{faster-rcnn}. In Faster R-CNN, the regions of interests (RoIs) are firstly generated by a Region Proposal Network (RPN). The generated RoIs are then refined and classified in the second stage. Different from two-stage detectors, one-stage detectors, including YOLO~\cite{redmon2016you} and SSD~\cite{liu2016ssd}, eliminate the region proposal generation step and directly predict bounding boxes and their associated classes. Our model is based on Faster R-CNN architecture. 


\begin{figure*}[ht!]
    \centering
    \includegraphics[width=\linewidth]{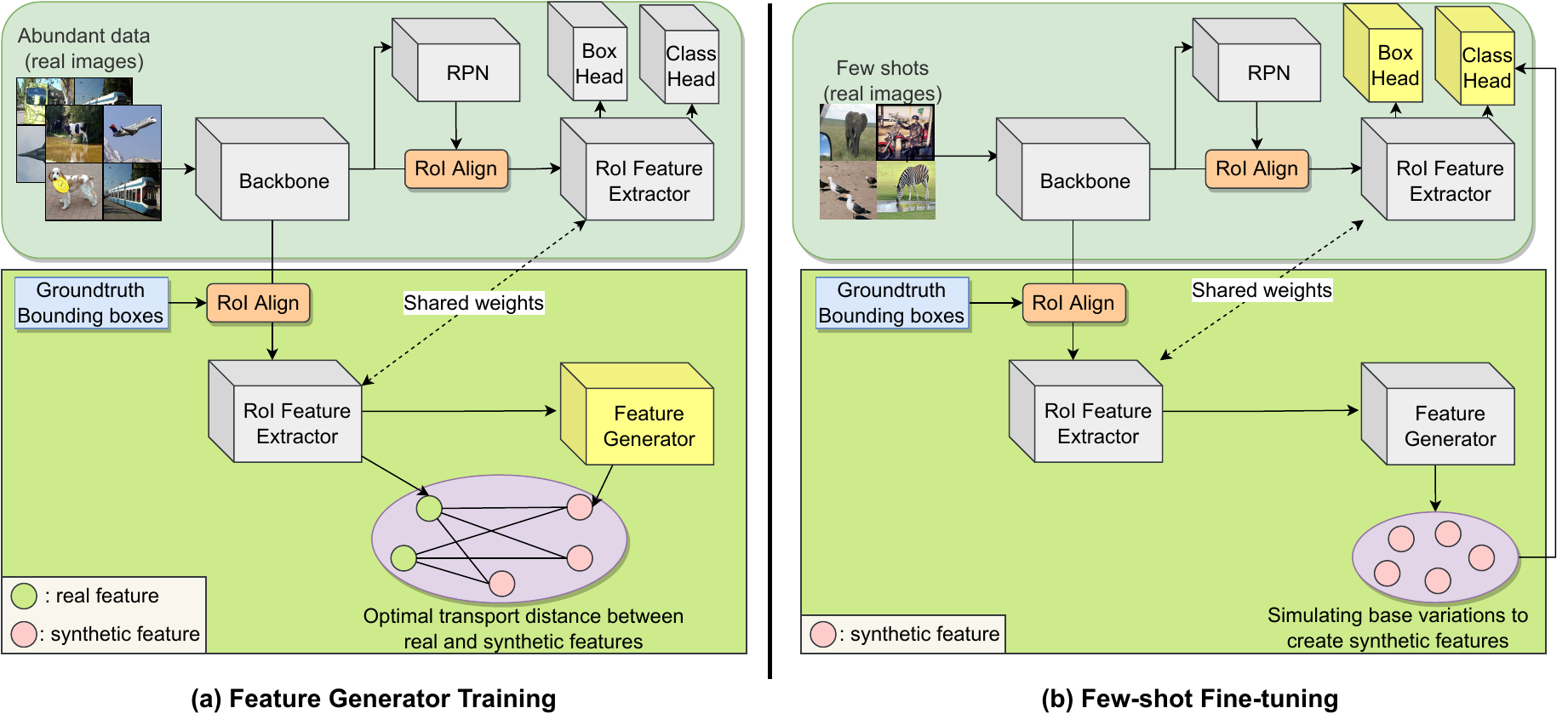}
    \caption{Overview of our approach. The top block is Faster R-CNN with backbone, RPN, RoI Feature Extractor, and RoI Heads modules. The bottom is our extended modules including RoI Feature Extractor and Generator $G$. (a) illustrates the feature generator training stage. We freeze Faster R-CNN and train generator $G$ on the abundant data of base classes. (b) presents applying pretrained feature generator $G$ to enhance the generalization of detector on the novel classes. The frozen generator creates synthetic features on few-shot samples of both base and novel classes to fine-tune the RoI Heads module. The gray rectangular box denotes the frozen module while the yellow rectangular box is the trainable module.}
    \label{fig:overview}
\vspace{-5mm}
\end{figure*}

\subsection{Few-Shot Object Detection} 

Few-Shot Object Detection (FSOD) aims to localize and classify the instances in the image with few training samples. FSOD approaches could be divided into two categories: single-branch and two-branch methods. Two-branch methods such as~\cite{meta-rcnn, yolo-reweighting} use the siamese network to extract the information from the query and the support images in parallel. Specifically, they use the similarity information between query and support features to classify the instances in the query image. The following works~\cite{max-margin, rpn-attention, hu2021dense, xiao2020few, han2021query, han2022meta} present the improvements such as applying attention mechanism in RPN, increasing the class margin, or aggregating multi-scale RoI features. Recent two-branch works~\cite{rpn-attention, hu2021dense} introduce feature attention in RPN and extract the proposals in multiple scales or relations. Other methods~\cite{max-margin, han2021query} leverage the intra- and inter-class to increase the feature discriminativeness. Other works exploiting the transformation~\cite{trans-int, wu2020multi}, transformer architecture~\cite{meta-detr, han2022few}, contrastive learning~\cite{sun2021fsce}, kernelized representations~\cite{zhang2022kernelized}, pseudo-labeling~\cite{kaul2022label} in FSOD. 
Single-branch methods~\cite{tfa, defrcn, zhang2021hallucination, zhu2021semantic} rely only on  query images. 
In~\cite{tfa} the authors present a simply two-stage fine-tuning approach (TFA)~\cite{tfa} which firstly trains the model on the base classes and then fine-tune few last layers on the novel classes. Other single-branch methods are proposed to deal with FSOD via extra text data~\cite{zhu2021semantic}, unlabeled image~\cite{khandelwal2021unit} or hallucination samples~\cite{zhang2021hallucination}. Recently, Decoupled Faster R-CNN (DeFRCN)~\cite{defrcn} achieves competitive results by stopping the backward gradients between modules and decoupling the multi-tasks in Faster R-CNN~\cite{faster-rcnn}. Lately, \cite{li2023disentangle} introduce D\&R  which is distilled from language models to enhance the generality of detectors while VFA \cite{han2023few} and apply variational autoencoders into FSOD. Another work \cite{lu2023breaking} aims to highlight the salient information to detect novel objects.


Different from previous FSOD methods, SFOT is designed to only leverage current dataset to synthesize new features. Without the support of external knowledge, SFOT can be integrated to boost the productivity of other methods.

\subsection{Optimal Transport} 

Optimal transport (OT) theory~\cite{villani2003topics, villani2009optimal} provides a powerful toolkit to compare probability distributions. However, OT has a high computational complexity (i.e., super cubic), which limits its ability to embed in many applications. Various proposals, e.g., entropic regularization~\cite{cuturi2013sinkhorn, benamou2015iterative}, $1$-dimensional slicing~\cite{rabin2011wasserstein}, tree-slicing~\cite{le2019tree}, low-rank approximation~\cite{scetbon2021low}, graph-based approach~\cite{le2022sobolev}, have been introduced in the literature to tackle this issue. OT~\cite{villani2009optimal} has been applied in different  problems~\cite{salimans2018improving, wu2021data}, including 3D shape recognition \cite{xu2019learning}, shape matching \cite{saleh2022bending}, label assignment \cite{ge2021ota}, and zero-shot learning \cite{wang2021zero}. 

\section{Proposed Method}

\subsection{Background}




We follow the standard settings of FSOD which are introduced in~\cite{tfa, defrcn}. {In FSOD, we have two disjoint sets consisting of base and novel classes, denoted $C_b$ and $C_n$, respectively}, i.e., $C_b \bigcap C_n = \O$. In the first stage, the detector is trained on base classes to build the concepts and considered as a $|C_b|$-class detector. In order to learn the new knowledge of novel classes while keeping the performance on base classes, the model is fine-tuned on a dataset in $C = C_b \cup C_n$. Note that in the fituning stage, there are only $K$-shot samples for each class in $C$. 

\subsection{Framework Overview}
Our proposed framework is presented in Figure~\ref{fig:overview}. There are two blocks: detector (i.e., Faster R-CNN~\cite{faster-rcnn}) and our proposed feature generator $G$. Figure \ref{fig:overview} (a) presents feature generator training. We keep freezing all layers of the pretrained $|C_b|$-class detector (including classifier) to 
guarantee the performance on $C_b$. In this stage, feature generator $G$ is optimized via the distribution distance between real features and synthetic features that it creates. The feature generator is then applied to the fine-tuning stage as shown in Figure \ref{fig:overview} (b). The feature generator module leverages base class distribution to simulate novel ones. Therefore, it could productively generate synthetic features in the novel class. 


\subsection{Conditional Feature Generator}

\begin{figure}[t!]
    \centering
    \includegraphics[width=\linewidth]{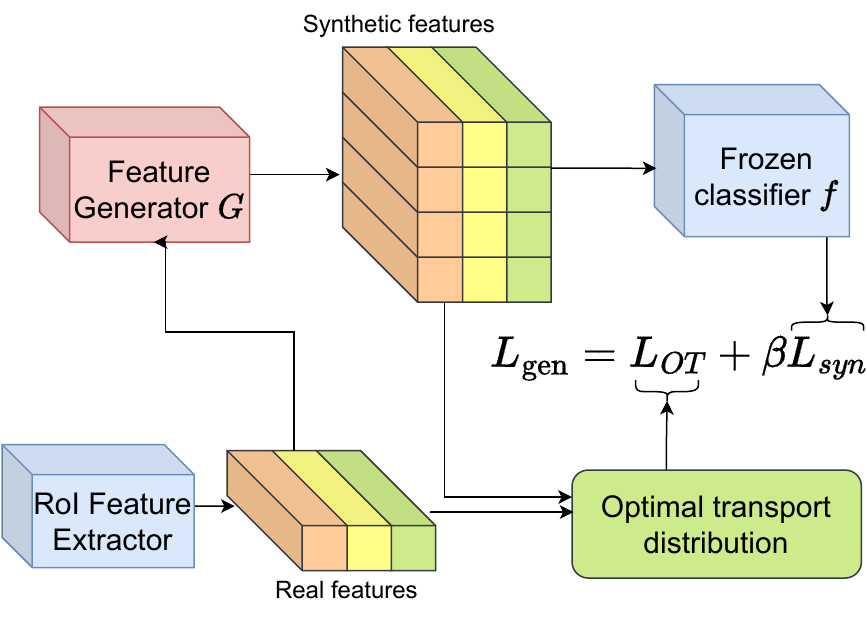}
    \caption{Feature Generator module. Feature Generator $G$ takes the input from RoI Feature Extractor and creates synthetic features. The generator loss function, which is used to update generator $G$, is based on optimal transport and frozen classifier $f$.}
    \label{fig:gen-train}
    \vspace{-5mm}
\end{figure}

With the aim to create a generator that leverages the variation on base classes $C_b$ to create synthetic features on novel classes $C_n$, we add a feature generator to pretrained detector (Figure \ref{fig:overview}) and train it on $C_b$. The feature generator module $G$ is illustrated in Figure~\ref{fig:gen-train}. Specifically, the real feature $x\in \mathbb{R}^d$ from RoI Feature Extractor is concatenated with a random vector $z \in \mathbb{R}^d$ which is sampled from a normal distribution, i.e., $z \sim \mathcal{N}(0,I_d)$, to create the synthetic feature $\hat{x}$.  

\begin{equation}
\label{eq:gen}
    \hat{x} = G([x; z]),
\end{equation}
where $[x; z]$ denotes the channel-wise concatenation of $x$ and $z$.  
Then, we use real features $x$ and synthetic features $\hat{x}$ to optimize generator $G$ via loss the function introduced in Equation (\ref{eq:gen_loss}). It is worth noting that 
 regarding the input of the RoI Feature Extractor, instead of using RoIs produced by the RPN that may not well capture objects, we use the groundtruth bounding boxes. This ensures the quality of the input for the generator training.

\subsection{Optimal Transport with Clustering Method}
\label{sec:ot_loss}



There are several distance measurements between real and synthetic feature sets,  e.g., KL-divergence, cross-entropy, and optimal transport (OT). 
However, point-wise distance only aligns individual synthetic, which is vulnerable to variations of samples, while OT provides an elegant approach to compare distributions even when supports are disjoint. Indeed, the OT lifts the ground metric between support data points (locally) to a distance between probability measures (globally). 
We utilize optimal transport as a loss function to optimize the generator $G$.


Suppose we have $N$ real inputs $\{x_n\}_{n=1}^N$ for the generator. By applying Equation (\ref{eq:gen}), we also obtain $M=N \times T$ synthetic features $\{\hat{x}_m\}_{m=1}^{M}$, where $T$ is the hyper-parameter to control the number of synthetic features.
However, when we generate abundant synthetic features to increase the diversity, this design also lifts computational costs including training time and complexity of the OT algorithm. Furthermore, there may be noisy samples in the generated samples, e.g., a sample follows the dog class but it is generated close to the car class. Directly using noisy samples for optimal transport will create ambiguity for the classifier and reduce the general performance.

To address the limitation, inspired by the hierarchical OT approach~\cite{yurochkin2019hierarchical}, we group synthetic features to $K$ centroids via a  clustering method, i.e., K-means++~\cite{arthur2006k}. Then, we compute the distribution distance between real features and $K$ centroids $\{e_k\}_{k=1}^K$. In this way, we not only decrease computational costs but also reduce the negative effects of noise samples. Additionally, this design allows us to take into account the number of elements of each cluster to create the distribution for synthetic features instead of using the uniform distribution.





Let $\mathbf{P}$ be the transportation plan matrix, we use the cosine distance for the ground cost matrix $\mathbf{C}_{nk}=1 - \frac{x_n^T e_k}{\norm{x_n}_2 \norm{e_k}_2}$. Let $\mathbb{S}^d$ be the probability simplex, i.e., $\mathbb{S}^{d-1} = \left\{a \in \mathbb{R}^{d} \mid a^T \mathbf{1}_d = 1, a \ge 0 \right\}$. The mass distributions are typically defined as $r =  \frac{1}{N} \mathbf{1}_N \in \mathbb{S}^{N-1}$ for real inputs $\{x_n\}_{n=1}^N$; and $c \in \mathbb{S}^{K-1}$ for $K$ centroids $\{e_k\}_{k=1}^K$. The $k$-th element of the mass distribution $c$ is defined as $c_k = \frac{\gamma_k}{\sum_{i=1}^K \gamma_i}$ where $\gamma_k$ is the number of elements of $k$-th clusters.
The OT loss between two distributions is defined as follows:


\begin{equation}\label{eq:OTloss}
    L_{OT}(r, c) = \min_{\mathbf{P} \in U(r,c)} \sum_{n=1}^N\sum_{k=1}^K \mathbf{P}_{nk}\mathbf{C}_{nk},
\end{equation}
where the transportation polytope $U(r, c) = \{\mathbf{P} \in \mathbb{R}^{N \times K}_+ | \mathbf{P}\mathbf{1}_K = r, \mathbf{P}^T\mathbf{1}_N = c\}$. 
The element $\mathbf{P}_{nk}$ indicates the probability of matching between real feature $x_n$ and synthetic centroid feature $e_k$. Moreover, we apply the entropic regularization for OT and use Sinkhorn algorithm \cite{cuturi2013sinkhorn} to further reduce the computational complexity from super cubic (for standard OT) into quadratic complexity to optimize $\mathbf{P}$. 




Additionally, we  utilize prior knowledge from the frozen classifier $f$ to update the generator. This guarantees that the synthetic vectors $\hat{x}$ must satisfy the pretrained classifier as the real vectors. In other words, it forces the synthetic vectors to retain the properties of the actual features. Specifically,  we freeze the classifier and use its output for the optimizing process of the generator $G$. 


Let $L_{syn}$ be the cross-entropy  between $\hat{y}$ (which is the label of $\hat{x}$) and the predicted class $f(\hat{x})$, i.e., $L_{syn} = -\sum_i \hat{y}_i \log(f(\hat{x}_i))$. 
The loss to train the generator $G$ is as follows:
\begin{equation}
\label{eq:gen_loss}
    L_{gen} = L_{OT} + \beta L_{syn},
\end{equation}
where $\beta$ is a hyper-parameter to balance between the two loss functions.

\subsection{Training Procedure}

The whole process includes three stages: base training, feature generator training, and few-shot fine-tuning. 
In the base training, we adopt the settings of~\cite{defrcn}. Specifically, we train the Faster R-CNN~\cite{faster-rcnn} with the scale gradient layers between modules. In this way, we can decouple tasks in Faster R-CNN to obtain a generalized $|C_b|$-class detector. After the base training, we freeze all the layers of $|C_b|$-class detector. Then, we add feature generator $G$ which is optimized via Equation (\ref{eq:gen_loss}).

In the fine-tuning stage, given a trained generator $G$, we aim to create synthetic samples like real counterparts on novel classes to enhance the generalization of methods. Similar to the TFA~\cite{tfa}, we only fine-tune the last layer of $|C|$-class detector. 
The new parameters are added to the classifier $f$ for novel classes. The new classifier $f$ is trained with both real and synthetic samples to capture the novel concepts. Specifically, to  train the detector, we feed the features $\hat{x}$ and $x$ of synthetic and real samples, respectively, into the classifier $f$ in the same updating process. Their predictions are used to update the classifier $f$ via the loss function:
\begin{equation}
\label{eq:loss-cls}
    L_{cls} = L_{real} + \alpha L_{syn},    
\end{equation}
where $L_{real}$ is the cross-entropy loss function for the real feature $x$. The hyper-parameter $\alpha$ is used to keep a balance between $L_{real}$ and $L_{syn}$.

\begin{table*}[ht]
\centering
\footnotesize
\addtolength{\tabcolsep}{-3.7pt}
\adjustbox{width=0.9\linewidth}{
\begin{tabular}{l|ccccc|ccccc|ccccc}
\toprule
\multirow{2}{*}{Method} & \multicolumn{5}{c|}{Novel Set 1} & \multicolumn{5}{c|}{Novel Set 2} & \multicolumn{5}{c}{Novel Set 3} \\ 
&  1     & 2     & 3    & 5    & 10   & 1     & 2     & 3    & 5    & 10   & 1     & 2     & 3    & 5    & 10   \\ \midrule
FSRW~\cite{yolo-reweighting}   & 14.8  & 15.5  & 26.7 & 33.9 & 47.2 & 15.7  & 15.3  & 22.7 & 30.1 & 40.5 & 21.3  & 25.6  & 28.4 & 42.8 & 45.9 \\ 
MetaDet~\cite{metadet} & 18.9 & 20.6 & 30.2 & 36.8 & 49.6 & 21.8 & 23.1 & 27.8 & 31.7 & 43.0 & 20.6 & 23.9 & 29.4 & 43.9 & 44.1 \\ 
Meta R-CNN~\cite{meta-rcnn}  & 19.9 & 25.5 & 35.0 & 45.7 & 51.5 & 10.4 & 19.4 & 29.6 & 34.8 & 45.4 & 14.3 & 18.2 & 27.5 & 41.2 & 48.1 \\ 
TFA w/ fc \cite{tfa} & {36.8} & {29.1} & {43.6} & {55.7} & {57.0} & {18.2} & {29.0} & {33.4} & {35.5} & {39.0} & {27.7} & {33.6} & {42.5} & {48.7} & {50.2}\\
TFA w/ cos \cite{tfa} & 39.8 & 36.1 & 44.7 & 55.7 & 56.0 & 23.5 & 26.9 & 34.1 & 35.1 & 39.1 & 30.8 & 34.8 & 42.8 & 49.5 & 49.8 \\ 
Xiao et al. \cite{xiao2020few} & 24.2 & 35.3 &  42.2 &  49.1 &  57.4 & 21.6 & 24.6 &  31.9 &  37.0 &  45.7 & 21.2 &  30.0 &  37.2 &  43.8 &  49.6 \\
MPSR \cite{wu2020multi} & 41.7 & 42.5 & 51.4 & 55.2 & 61.8 & 24.4 & 29.3 & 39.2 & 39.9 & 47.8 & 35.6 & 41.8 & 42.3 & 48.0 & 49.7 \\ 
Fan et al. \cite{rpn-attention} & 37.8 & 43.6 & 51.6 & 56.5 & 58.6    & 22.5 & 30.6 & 40.7 & 43.1 & 47.6    & 31.0 & 37.9 & 43.7 & 51.3 & 49.8\\ 
SRR-FSD \cite{zhu2021semantic}  & {47.8} & 50.5 & 51.3 & 55.2 & 56.8    & 32.5 & 35.3 & 39.1 & 40.8 & 43.8    & 40.1 & 41.5 & 44.3 & 46.9 & 46.4 \\
TFA + Halluc \cite{zhang2021hallucination} & 45.1 & 44.0 & 44.7 & 55.0 & 55.9   & 23.2 & 27.5 & 35.1 & 34.9 & 39.0   & 30.5 & 35.1 & 41.4 & 49.0 & 49.3 \\
CoRPNs + Halluc \cite{zhang2021hallucination} & 47.0 & 44.9 & 46.5 & 54.7 & 54.7   & 26.3 & 31.8 & 37.4 & 37.4 & 41.2   & 40.4 & 42.1 & 43.3 & 51.4 & 49.6 \\
FSCE \cite{sun2021fsce} &  44.2 & 43.8 & 51.4 & 61.9 & 63.4    & 27.3 & 29.5 & 43.5 & 44.2 & 50.2    & 37.2 & 41.9 & 47.5 & 54.6 & 58.5 \\
Meta Faster R-CNN \cite{han2022meta} & 43.0 & 54.5 & 60.6 & 66.1 & 65.4   & 27.7 & 35.5 & \textbf{46.1} & 47.8 & 51.4   & 40.6 & 46.4 & 53.4 & 59.9 & 58.6\\
KFSOD\cite{zhang2022kernelized} & 44.6 & - & 54.4 & 60.9 & 65.8 & \textbf{37.8} & - & 43.1 & 48.1 & 50.4 & 34.8 & - & 44.1 & 52.7 & 53.9 \\
D\&R \cite{li2023disentangle} & 41.0  & 51.7  & 55.7  & 61.8  & 65.4  & 30.7  & 39.0  & 42.5  & 46.6  & 51.7  & 37.9  & 47.1  & 51.7  & 56.8  & 59.5 \\
VFA \cite{han2023few} & 47.4 & 54.4 & 58.5 & 64.5 & 66.5 & 33.7 & 38.2 & 43.5 & 48.3 & 52.4 & \textbf{43.8} & 48.9 & 53.3 & 58.1 & 60.0\\

\midrule
DeFRCN \cite{defrcn}  & 40.2 & 53.6 & 58.2 & 63.6 & 66.5 & 29.5 & 39.7 & 43.4 & 48.1 & 52.8 & 35.0 & 38.3 & 52.9 & 57.7 & 60.8 \\
SFOT (Ours) & \textbf{47.9} & \textbf{60.4} & \textbf{62.7} & \textbf{67.3} & \textbf{69.1} & 32.4 & \textbf{41.2} & 45.7 & \textbf{50.2} & \textbf{54.0} & 43.5 & \textbf{54.1} & \textbf{56.9} & \textbf{60.6} & \textbf{62.5} \\
\bottomrule
\end{tabular}}
\caption{{Few-shot object detection performance (nAP50) on the PASCAL VOC dataset.} Note that we report the averaged result over 10 runs. The best performance is marked in \textbf{boldface}.} 
\label{tab:main_voc}
\vspace{-3mm}
\end{table*}

\section{Experimental Results \& Discussion}
\subsection{Experimental Settings}
\paragraph{Datasets.}
We evaluate our method on two widely-used few-shot object detection benchmarks: PASCAL VOC \cite{everingham2010pascal, everingham2015pascal}  and MS COCO \cite{lin2014microsoft}. To fairly compare with other methods, we follow the setup from DeFRCN~\cite{defrcn}. For PASCAL VOC, the dataset is considered a standard benchmark of few-shot object detection. The whole dataset consists of 20 classes and is partitioned into three different sets. There are five classes for the novel set and 15 remaining ones for the base set. We use the same novel/base classes for a fair comparison. Each set holds $K \in \{1, 2, 3, 5, 10\}$ instances of both base and novel classes. Regarding MS COCO, the dataset is a more challenging benchmark of FSOD. In few-shot object detection setting, 80 classes of MS COCO are divided into 60 classes for the base set and 20 classes for the novel set.

\paragraph{Evaluation Setting.} We adopt the evaluation protocols for generalized few-shot object detection of DeFRCN~\cite{defrcn} to consider the effectiveness of our approach. This provides a comprehensive evaluation of the performance of the few-shot detector in both base and novel classes. Following the standard benchmark~\cite{zhang2021hallucination, defrcn, han2022meta}, we report AP50 on all three sets; and AP and AP75 for more detailed discussion in novel set 1. For MS COCO, we report AP50. The main results are averaged over 10 repeated runs with different random seeds.


\subsection{Implementation Details}

\paragraph{Hyper-parameter Settings.} In the first base training phase, we use the SGD optimizer with the hyper-parameters mentioned in  DeFRCN~\cite{defrcn}. In the feature generator training, we freeze all the layers of $|C_b|$-class detector. We use the learning rate of 0.02 and the batch size of 64 and set $\beta=0.1, T=16, S=128$. On PASCAL VOC, we train our generator $G$ with $1000$ iterations and decay the learning rate by 0.1 ratios at $250$ and $750$ iterations. On MS COCO, generator $G$ is trained with $1500$ iterations and the learning rate decay at $300$ and $1200$ iterations. In fine-tuning stage, we set $\alpha=0.01$ in Equation~(\ref{eq:loss-cls}), $T=512$ and $K=32$. We only train two last layers like TFA~\cite{tfa} ( classifier and box-regressor), other training hyper-parameters are set as the default of baseline. We also adopt the evaluation procedure mentioned in ~\cite{defrcn} to fairly compare with existing methods.

\paragraph{Model Architecture.} Our detector adopts Faster R-CNN~\cite{faster-rcnn} settings like DeFRCN~\cite{defrcn}. In the fine-tuning stage, we use cosine classifier~\cite{tfa} to be consistent with the optimal transport problem, which is based on the cosine distance. The feature generator $G$ architecture consists of fully-connected (FC) layers with ReLU/Leaky ReLU.

\begin{table}[!t]
\centering
\addtolength{\tabcolsep}{-2.5pt}
\adjustbox{width=0.85\linewidth}{
\footnotesize
\begin{tabular}{l|lll|lll}
\toprule
\multicolumn{1}{c}{Method}& \multicolumn{1}{|c}{bAP} & \multicolumn{1}{c}{bAP50} & \multicolumn{1}{c|}{bAP75} & \multicolumn{1}{c}{nAP} & \multicolumn{1}{c}{nAP50} & \multicolumn{1}{c}{nAP75} \\\midrule
Base model & 54.6 & 80.5	& 60.1 & - & - & -\\\midrule

\multicolumn{7}{c}{1-shot} \\\midrule
DeFRCN & 48.1 & 75.2 & 53.0 & 23.0 & 41.6 & 23.1 \\
SFOT & 50.4 & 78.0 & 55.6 & 26.1 & 47.9 & 25.2 \\\midrule
\multicolumn{7}{c}{3-shot} \\\midrule
DeFRCN & 48.8 & 76.0 & 53.5 & 34.4 & 59.6 & 34.6 \\
SFOT & 51.2 & 78.2 & 56.6 & 35.9 & 62.7 & 35.7 \\\midrule
\multicolumn{7}{c}{10-shot} \\\midrule
DeFRCN & 49.8 & 76.4 & 54.8 & 39.2 & 66.0 & 40.7 \\
SFOT & 51.6 & 78.1 & 57.0 & 41.2 & 69.1 & 42.7\\\bottomrule 
\end{tabular}}
\caption{The performance of base classes and novel classes on Novel Set 1 of VOC dataset. We re-implement the results of our baseline DeFRCN~\cite{defrcn} and report the averaged result over 10 runs.} 
\vspace{-4mm}
\label{tab:base-classes-voc}
\end{table}

\begin{table}[t!]
\centering
\footnotesize
\addtolength{\tabcolsep}{-2.5pt}
\adjustbox{width=\linewidth}{
\begin{tabular}{l|ccc}
\toprule
Method & 1 & 2 & 10 \\ \midrule
FSRW \small{~\cite{yolo-reweighting}}          & -             & -             & 12.3          \\
MetaDet \small{~\cite{metadet}}                        & -             & -             & 14.6          \\
Meta R-CNN \small{~\cite{meta-rcnn}}           & -             & -             & 19.1          \\
TFA w/ fc \small{\cite{tfa}}                          & 5.7           & 8.5           & 19.2          \\
TFA w/ cos \small{\cite{tfa}}                         & 5.8           & 8.3           & 19.1          \\
Xiao et al. \small{\cite{xiao2020few}}                & 8.9           & 13.3          & 25.6          \\
MPSR \small{\cite{wu2020multi}}                       & 4.1           & 6.3           & 17.9          \\
Fan et al. \small{\cite{rpn-attention}}               & 9.1           & 14.0          & 20.7          \\
SRR-FSD \small{\cite{zhu2021semantic}}                & -             & -             & 23.0          \\
TFA + Halluc \small{\cite{zhang2021hallucination}}    & 7.5           & 9.9           & -             \\
CoRPNs + Halluc \small{\cite{zhang2021hallucination}} & 6.5           & 9.0           & -             \\
FSCE \small{\cite{sun2021fsce}}                       & -             & -             & -             \\
Meta Faster R-CNN \small{\cite{han2022meta}}          & 10.7          & 16.3          & 25.7          \\ \hline
DeFRCN \small{\cite{defrcn}}                        & 9.5           & 16.3          & 29.6          \\
SFOT (Ours)                                   & \textbf{13.0} & \textbf{19.3} & \textbf{29.8} \\ \hline
\end{tabular}}
\caption{Few-shot object detection performance (nAP50) on MS COCO dataset. Note that we report the averaged result over 10 runs. The best performance is marked in boldface.} \vspace{-5mm}
\label{tab:main_coco}
\end{table}

\subsection{Comparison with SOTA Methods}

\paragraph{Results on PASCAL VOC.} We report the peformance in terms of AP50 in Table \ref{tab:main_voc}. Generally, our proposed method outperforms the existing methods in all FSOD settings of PASCAL VOC. Particularly, we averagely improve 4.1\%, 2.2\%, and 4.2\% in novel set 1, 2, and 3, respectively. 


In novel set 1,  SFOT outperforms all baselines with a large margin. We particularly enhance the generalization of $|C|$-class detector in the context of extremely scarce labeled data, achieving 47.7\% (improved 6.3\%) in 1-shot and 60.4\% in 2-shot (improved 5.1\%). In the recent work, Zhang\textit{ et al.} \cite{zhang2021hallucination} also generate synthetic samples, but only improve the performance in the context of extremely scarce data. The reason is that they only use the pretrained classifier which is fine-tuned on novel classes with few samples to create synthetic features. Therefore, their models are overfitting novel classes, and there is no significant improvement when there exist more training data. Meanwhile, we improve the generalization of FSOD detectors (averaged 2.8\% for 2+ shots in all three sets) when more training labeled data are available. These results demonstrate the effectiveness of our generator in few-shot object detection, which can leverage the variations on base classes to productively create synthetic features on novel classes. Therefore, it could enhance the performance of the FSOD models in novel classes. This situation also occurs during experiments in novel set 3 of PASCAL VOC, namely, surpassing 1.4\% -- 7.9\% when compared to DeFRCN~\cite{defrcn}.

We additionally show the effectiveness of our method SFOT in comparison with the DeFRCN baseline in Table \ref{tab:base-classes-voc}. We present the performance in three common metrics including AP, AP50, and a stricter metric - AP75 in both novel and base classes. The results show that in addition to improving the results in novel classes, SFOT also maintains the good performance of methods in base classes. Specifically, in terms of AP, we get better results (improved 1.0\% -- 3.1\%) in novel classes and especially improve by about 2\% performance in base classes. Besides, our method could recreate the former distribution better when there exist more training data. Note that the ineffective synthetic features could harm both novel and base classes. The previous work \cite{zhang2021hallucination} can create effective features for the novel classes, but it also reduces the performance of baseline models on base classes. Therefore, SFOT is better than the previous method for both base classes $C_b$ and novel classes $C_n$ via the given empirical evidence. The reason is that our generator captures the distribution of classes instead of each individual in a specific class to effectively generate synthetic features.

\begin{figure*}[!t]
    \centering
    \includegraphics[width=\linewidth]{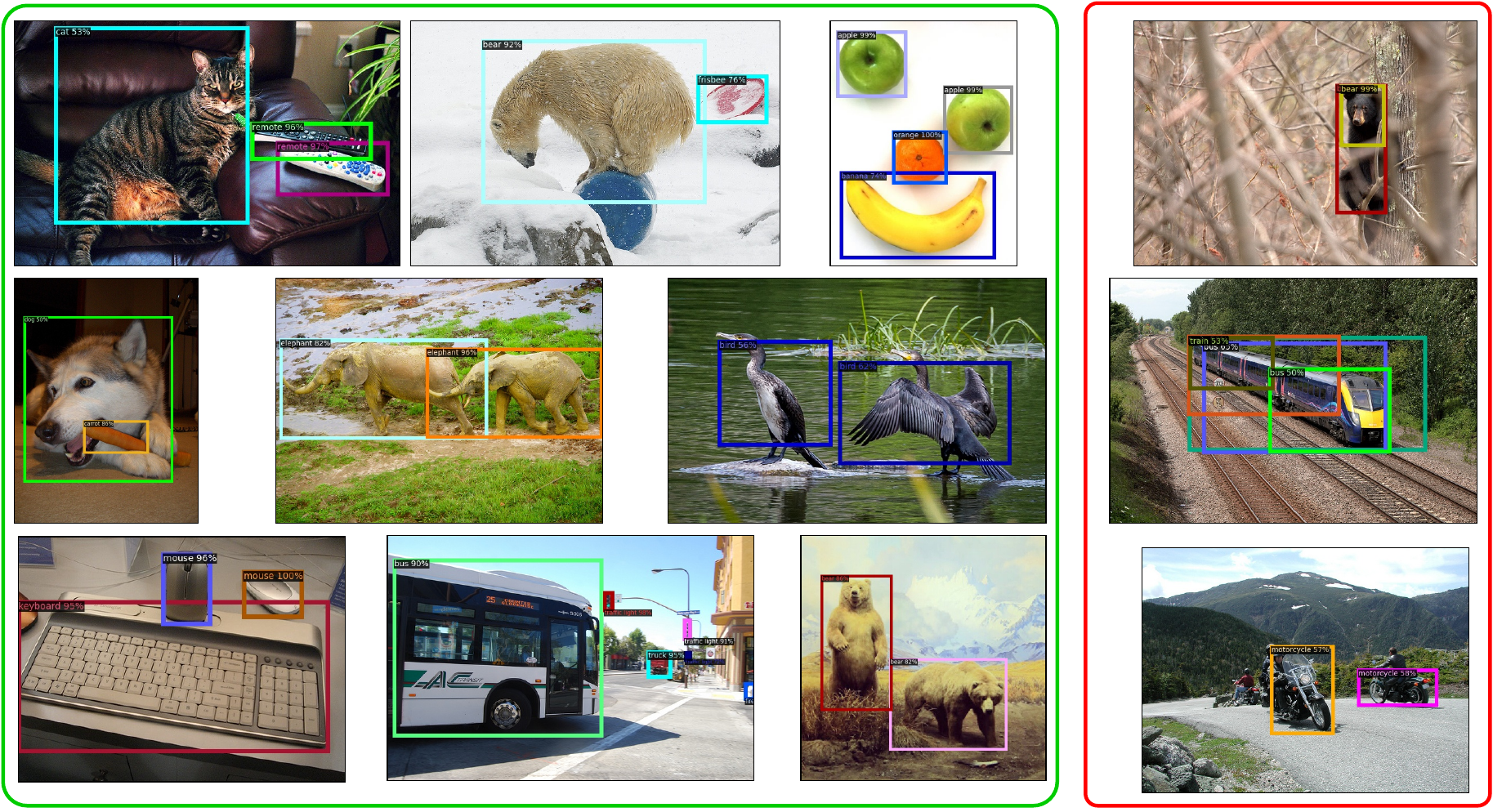}    \caption{Illustration results of SFOT on 2-shot of MS COCO. The green round rectangle demonstrates good cases of our approach while the red round rectangle illustrates failure cases.}
\label{fig:visualization}
\end{figure*}


\begin{table}[!t]
\adjustbox{width=\linewidth}{
\begin{tabular}{l|ccc|ccc}
\toprule
Method & bAPs & bAPm & bAPl & nAPs & nAPm & nAPl \\\midrule
DeFRCN & 14.1 & 34.7 & 44.0 & 1.1  & 3.6  & 8.7  \\
SFOT   & \textbf{17.2} & \textbf{37.8 }& \textbf{47.2} & \textbf{2.2}  & \textbf{6.1}  & \textbf{9.6} \\\bottomrule
\end{tabular}}
\caption{Additional results following the size of the objects of base and novel classes in MS COCO.}
\label{tab:coco_base_novel_size}
\vspace{-6mm}
\end{table}
\begin{table}[!b]
\vspace{-5mm}
\adjustbox{width=\linewidth}{
\begin{tabular}{l|cccccc|c}
\toprule
Method &  Meta R-CNN & MPSR & SRR-FSD & DeFRCN & SFOT \\\midrule
mAP50  &  37.4       & 42.3 & 44.5    & 55.9   & \textbf{60.0}       \\
\bottomrule
\end{tabular}}
\caption{10-shot cross-domain performance in  VOC.}
\label{tab:coco2voc}
\end{table}

\paragraph{Results on MS COCO.} We conduct the experiment on MS COCO dataset and report the novel performance (nAP50) of all FSOD methods in Table \ref{tab:main_coco}. As shown in the table, our SFOT surpasses all baselines to achieve the state-of-the-art performance for all settings, namely, 1-shot, 2-shot, and 10-shot. In particular, our method significantly outperforms the most recent method, Meta Faster R-CNN~\cite{han2022meta}, by a large margin. In particular, when data is extremely scarce (i.e., 1-, 2-shot), our method drastically improves the results of the baseline DeFRCN~\cite{defrcn}.

In addition, we provide the evaluations on the size of objects in Table \ref{tab:coco_base_novel_size}. The results in 1-shot show that SFOT surpasses comprehensively in comparison with DeFRCN \cite{defrcn} in the context of limited available data. As can be seen from the Table, our approach improves DeFRCN baseline not only on large-, medium- and small-size objects on novel classes but also on base classes as well. Specifically,  our model get averaged results of 17.2\% bAPs, 37.8\% bAPm, and 47.2\% APl, with an improvement of 3.1\%, 3.1\%, and 3.2\% compared to DeFRCN~\cite{defrcn}. One possible reason is that our generator can learn the concept of objects of various sizes in base classes via OT loss function mentioned in Equation (\ref{eq:gen_loss}). Therefore, the module can leverage the size information to create synthetic features for novel and base objects during the fine-tuning stage. 


\paragraph{MS COCO to PASCAL VOC.} Following the previous method~\cite{zhu2021semantic,defrcn}, we  conduct experiments on the cross-domain setting.  SFOT achieves the best performance with 60.0\% in Table~\ref{tab:coco2voc}, demonstrating our method has better generality in the cross-domain situations.

\paragraph{Qualitative Results.} We visualize the detection results of 2-shot of MS COCO in Figure~\ref{fig:visualization}. As seen in the figure, our proposed method can well detect various objects in various sizes of both novel and base classes. However, SFOT has several failure cases as well. There exist overlapping bounding boxes in the same object. Especially, the detection of persons is often inaccurate due to the large variety, and people are often obscured by other objects. 




\begin{table}[!t]
\adjustbox{width=\linewidth}{
\begin{subtable}{0.5\linewidth}
\resizebox{0.97\textwidth}{!}{%
\begin{tabular}{l|ccc}
Method & nAP & nAP50 & nAP75 \\\midrule
FC w/ cos & 32.5 & 57.7 & 32.1 \\
{SFOT w/cos} & \textbf{33.8} & \textbf{60.4} & \textbf{33.1} \\ \bottomrule               
\end{tabular}
}
\caption{Cosine classifier} 
\label{tab:ab_cosine}
\vspace{-2mm}
\end{subtable}
\begin{subtable}{0.5\linewidth}
\adjustbox{width=0.97\linewidth}{\begin{tabular}{l|lll}
\toprule
{Method} 
 & AP & AP50 & AP75 \\\midrule
L2 & 32.3 & 57.6 & 32.0  \\
KL & 32.9 & 58.6 & 32.8 \\
Our OT & \textbf{33.8} & \textbf{60.4} & \textbf{33.1}  \\\bottomrule
\end{tabular}}
\caption{Generation loss functions.}
\vspace{-2mm}
\label{tab:ablation_voc_loss_gen}
\end{subtable}
}
\caption{Ablation experiments about the cosine classifier and generation loss function on 2-shot of Novel Set 1 in VOC.}
\vspace{-5mm}
\end{table}


\subsection{Ablation Study}

\paragraph{Efficient for our feature generator with cosine classifier.} We further evaluate the cosine classifier with and without our generator in Table \ref{tab:ab_cosine}. The results show that our feature generator, which is based on the cosine distance, achieves significant performance on  PASCAL VOC. Specifically, our method improves by around 3\% in AP50 of novel classes and it also boosts the performance of base classes. 


\paragraph{The impact of $K$.} we report the impact of hyper-parameters $K$ of our method in Table \ref{tab:ab_KT} where we utilize the same $T=512$ for both generator training and fine-tuning stage to conduct novel AP. We observe that nAP increases when $K$ rises and peaks at 26.08, with $K=32$.



\paragraph{Generator loss function.} Our method could use different loss functions such as L2 or KL to match the distribution between real features and synthetic features. In Table \ref{tab:ablation_voc_loss_gen}, we provide ablation results about generation loss functions and it shows that our OT achieves outstanding performance.

\paragraph{Universal model.} Our method could apply to other FSOD methods. For instance, we combine our method with TFA~\cite{tfa} and report results in Table
\ref{tab:tfa_performance}. SFOT can improve TFA by about 2\%.


\begin{table}[t]
\begin{tabular}{l|cccccc}
\toprule
K   & 4     & 8     & 16    & 32    & 64    & 128   \\\midrule
nAP & 25.05 & 25.04 & 25.26 & \textbf{26.08} & 25.42 & 25.43 \\\bottomrule
\end{tabular}
\caption{Ablation hyper-parameters $K$.  We report novel performance of 1-shot of Novel Set 1 in PASCAL VOC.}
\label{tab:ab_KT}
\vspace{-3mm}
\end{table}

\begin{table}[bt!]
\centering
\adjustbox{width=0.7\linewidth}{
\begin{tabular}{l|cc}
\toprule
Method   & 1-shot  & 5-shot \\\midrule
TFA/TFA+SFOT & 37.3/\textbf{39.9} & 43.9/\textbf{45.3} \\\bottomrule
\end{tabular}}
\caption{nAP50 of TFA,  TFA+SFOT on Novel Set 1 in VOC.}
\label{tab:tfa_performance}
\vspace{-6mm}
\end{table}

\section{Conclusion}

In this paper, we present a new method to tackle FSOD via generating synthetic samples. In particular, we estimate the distribution matching between real and synthetic samples via tackling the optimal transport problem with clustering method to effectively train the feature generator. Our model simulates the base distribution to beneficially create synthetic features and enhance the generalization of detectors when training data is insufficient. Our model simulates the base distribution to beneficially create synthetic features and enhance the generalization of detectors when training data is insufficient, which is demonstrated by empirical evidence on two benchmarks.



\bibliography{aaai24}

\end{document}